\documentclass[11pt, a4paper, logo, twocolumn]{deepmind}

\pdftrailerid{redacted}

\usepackage{dsfont}
\usepackage{dm-colors}

\usepackage[authoryear, sort&compress, round]{natbib} 

\usepackage{subfigure}

\usepackage{cleveref}

\graphicspath{{}}

\title{Is Curiosity All You Need? On the Utility of Emergent Behaviours from Curious Exploration}

\correspondingauthor{ogroth@robots.ox.ac.uk}


\reportnumber{} 

\renewcommand{\today}{2021-09-17}

\newcommand{\selmourl}{\url{https://deepmind.com/research/publications/2021/Is-Curiosity-All-You-Need-On-the-Utility-of-Emergent-Behaviours-from-Curious-Exploration}}

\author[1,2]{Oliver Groth}
\author[1]{Markus Wulfmeier}
\author[1]{Giulia Vezzani}
\author[1,3]{Vibhavari Dasagi}
\author[1]{Tim Hertweck}
\author[1]{Roland Hafner}
\author[1]{Nicolas Heess}
\author[1]{Martin Riedmiller}

\affil[1]{DeepMind}
\affil[2]{University of Oxford}
\affil[3]{Queensland University of Technology}

\begin{abstract}
Curiosity-based reward schemes can present powerful exploration mechanisms which facilitate the discovery of solutions for complex, sparse or long-horizon tasks.
However, as the agent learns to reach previously unexplored spaces and the objective adapts to reward new areas, many behaviours emerge only to disappear due to being overwritten by the constantly shifting objective. 
We argue that merely using curiosity for fast environment exploration or as a bonus reward for a specific task does not harness the full potential of this technique and misses useful skills.
Instead, we propose to shift the focus towards retaining the behaviours which emerge \emph{during} curiosity-based learning.
We posit that these self-discovered behaviours serve as valuable skills in an agent's repertoire to solve related tasks. Our experiments demonstrate the continuous shift in behaviour throughout training and the benefits of a simple policy snapshot method to reuse discovered behaviour for transfer tasks.
\end{abstract}

\begin{document}

\maketitle

\section{Introduction}
Intrinsic motivation~\citep{schmidhuber1991curiosity_boredom,oudeyer2007intrinsic_motivation,oudeyer2009intrinsic,baranes2009riac,schmidhuber2010formal_creativity} can be a powerful concept to endow an agent with an automated mechanism to continuously explore its environment in the absence of task information.
One common way to implement intrinsic motivation is to train a predictive model alongside the agent's policy and use the model's prediction error as a reward signal for the agent encouraging the exploration of previously unfamiliar transitions in the environment - a method also known as \emph{curiosity learning}~\citep{pathak2017curiosity}.
Curiosity-esque reward schemes have been used in different ways to facilitate exploration in sparse tasks~\citep{houthooft2016vime,burda2018rnd} or pre-train policy networks before fine-tuning them on difficult downstream tasks~\citep{sekar2020planning_to_explore}.
In environments where the main task objective is highly correlated with thorough exploration, curiosity-based approaches have also been shown to solve the main task without any additional reward signal~\citep{burda2018curiosity_study}.

However, in environments with multiple possible tasks -- e.g. in manipulation scenarios where objects could be interacted with or re-arranged in different ways -- not only the final behaviour of a curious exploration run might be of interest, but intermediate behaviours can correlate with solutions to different tasks.
Naturally, the constantly changing curiosity objective leads to the emergence of diverse behaviours during training -- much akin to the learning process of infants which develop  useful skills by playing~\citep{haber2018emergence}.
Yet, only a fraction of this diversity is ultimately retained in a downstream, task-specific policy -- which might also be biased towards exploration as a side-effect -- or it is even completely overwritten after fine-tuning.
This problem of \emph{catastrophic forgetting}~\citep{mccloskey1989catastrophic} is well-known for any neural network which operates under a shifting data distribution.
However, the intermediate behaviours which emerge and disappear \emph{during} learning based on curiosity can be of relevance for different tasks of interest.
If we were able to extract and leverage emergent behaviour, we could turn the process of exploration from a service for task-driven reinforcement learning into a rich continual learning setup in its own right~\citep{hadsell2020embracing}.

\begin{figure*}[t]
\vskip -0.2in
\begin{center}
\centerline{\includegraphics[width=\textwidth]{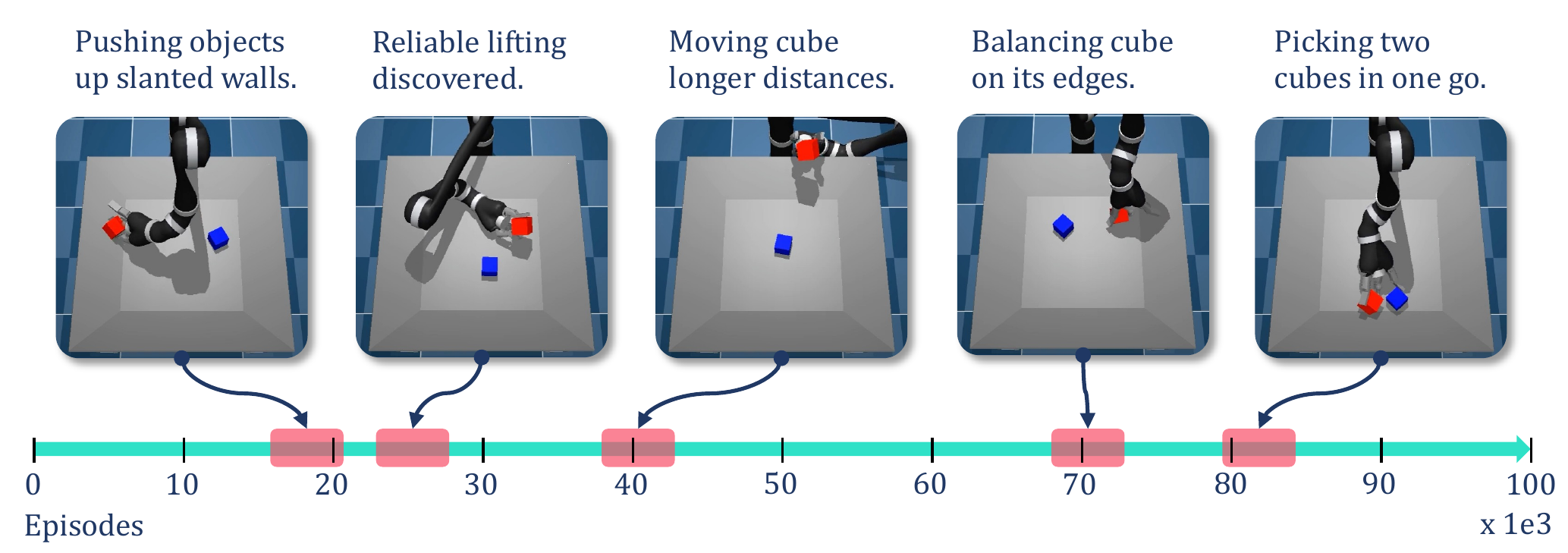}}
\centerline{\includegraphics[width=\textwidth]{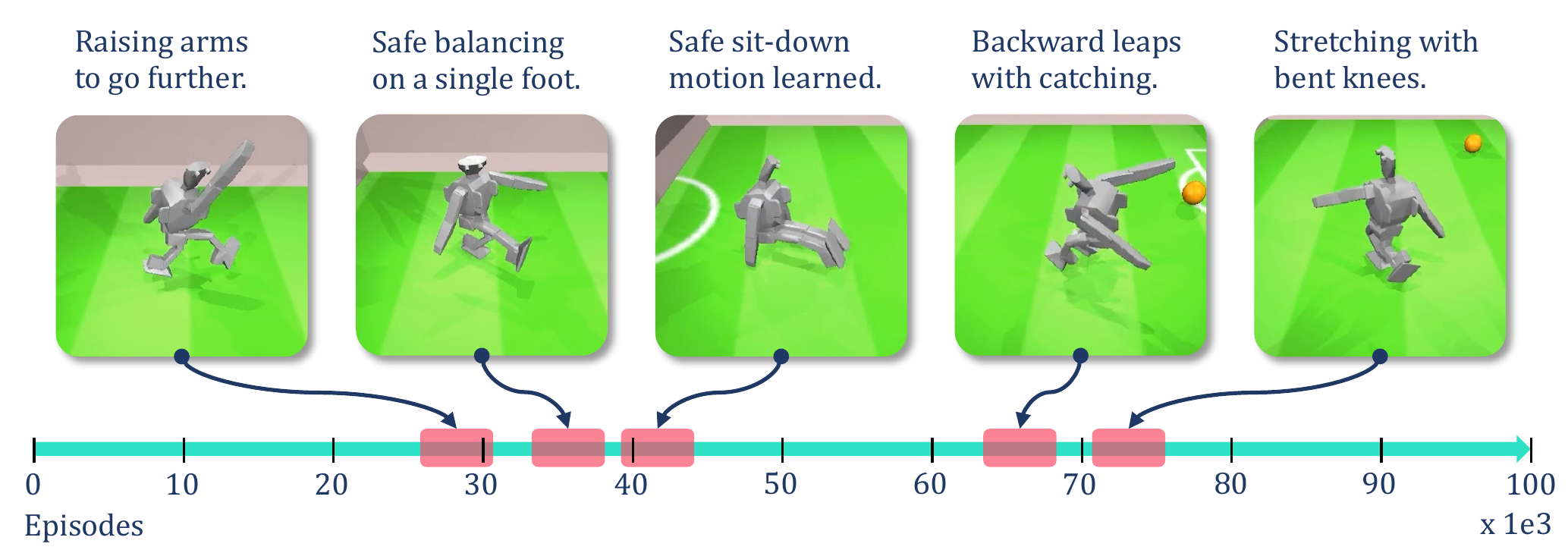}}
\caption{
Two example timelines depicting the emergence of behaviour while pursuing a curiosity objective on a 9-DoF JACO arm (top) and on a 20-DoF OP3 humanoid robot (bottom).
Each timeline represents the evolution of behaviour from a single random seed on a single simulated actor.
At each point in time, the agent exhibits a single behaviour which slowly evolves over time as the curiosity objective changes.
A detailed description of the emergent behaviour in this experiment is provided in~\cref{sec:exp-emergence} and corresponding quantitative results are shown in~\cref{fig:timeline-quantitative-jaco,fig:timeline-quantitative-op3}.
The corresponding videos can be found at: \selmourl.
}
\label{fig:timeline-qualitative}
\end{center}
\vskip -0.2in
\end{figure*}

Despite technical challenges, the discovery of self-induced curricula of skills holds a tantalising prospect for an agent's ability to solve broad sets of long-horizon tasks when it is able to draw upon potentially useful skills.
For instance, a robotic arm which has already discovered how to reach, grasp and lift objects in its workspace has a much easier time exploring and learning a policy to stack objects later on, if it recombined its previously acquired skills.
Recent works~\citep{riedmiller2018sacx,hertweck2020ssi,wulfmeier2020representation} have studied the influence of specific tasks across a spectrum of manual engineering effort like in the stacking example on the learning success of complex manipulation policies.
A curiosity-based approach could further reduce the effort of designing a curriculum of tasks and reward functions with a set of self-discovered skills.

In this paper, we study behaviour which emerges based on a curiosity objective in two continuous control settings: manipulation and locomotion.
In contrast to prior work in this domain, we implement curiosity-based exploration in an off-policy learning setting which improves upon on-policy implementations in terms of data-efficiency and presumably increases the diversity of emerging behaviours.
Furthermore, we look at the utilisation of the self-discovered behaviour for learning new downstream tasks.
In particular, we find that even a na\"ive baseline which treats snapshots of a curiosity policy as fixed exploration skills in a hierarchical learning setup can perform commensurately with a hand-designed curriculum learning setup.
Our findings suggest that the identification and exploitation of self-discovered behaviours can be a fruitful avenue for future work in the learning of complex robotics tasks.

In summary, we make the following two contributions:
First, we introduce \emph{SelMo}, an off-policy realisation of a self-motivated, curiosity-based method for exploration which is applied to two robotic manipulation and locomotion domains in simulation.
We show that even in those complex, 3D environments, meaningful and diverse behaviour emerges solely based on the optimisation of the curiosity objective.
Second, we propose to extend the focus in the application of curiosity learning towards the identification and retention of emerging intermediate behaviours and support this conjecture with a baseline experiment which uses self-discovered behaviours as auxiliary skills in a hierarchical reinforcement learning setup.

\begin{figure*}[ht]
\begin{center}
\centerline{\includegraphics[width=\textwidth]{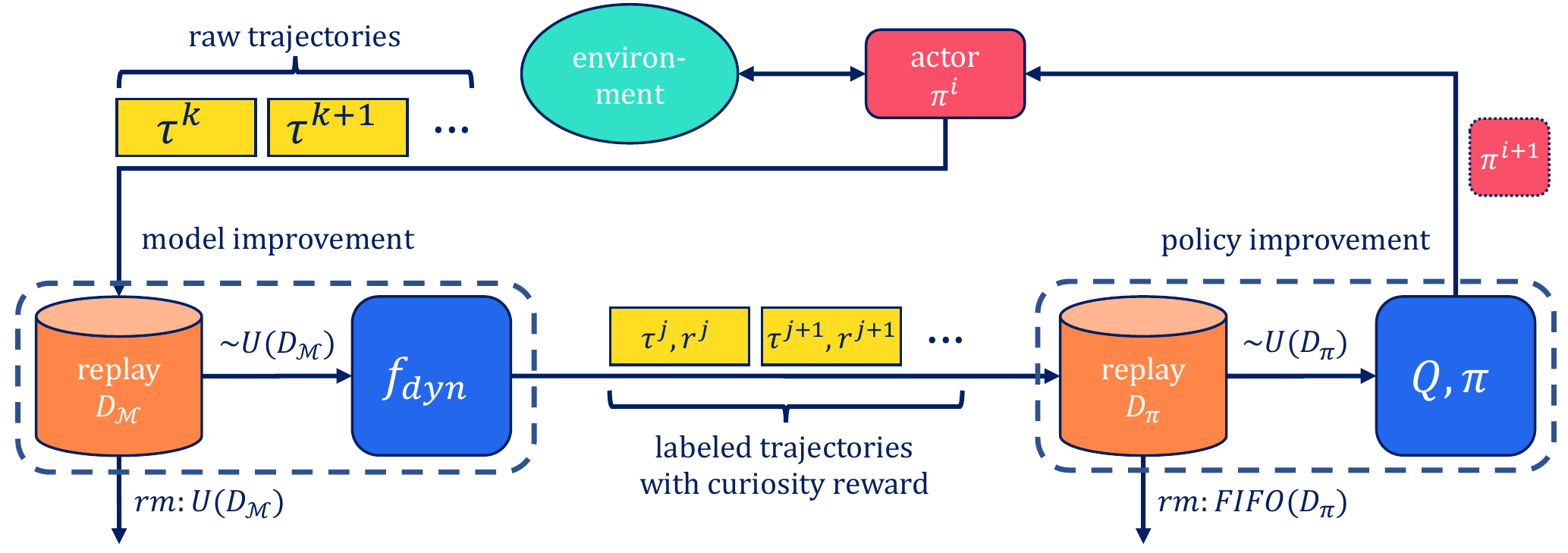}}
\caption{
An overview over the SelMo system architecture.
The agent collects trajectories $\tau^k, \tau^{k+1}, \dots$ in the environment using its current policy $\pi^i$ and stores it in a model replay buffer $D_\mathcal{M}$.
When $D_\mathcal{M}$ is full, trajectories are replaced with a uniform removal strategy.
The dynamics model $f_{dyn}$ samples uniformly from this buffer and updates its parameters for forward prediction using \emph{stochastic gradient descent} (SGD).
The sampled trajectories $\tau^j, \tau^{j+1}, \dots$ are then assigned a curiosity reward $r^j, r^{j+1}, \dots$ scaled by their respective prediction error under the current $f_{dyn}^{(j)}$.
The labeled trajectories are passed on to the policy replay buffer $D_\pi$ which runs a $FIFO$ removal strategy.
\emph{Maximum a posteriori policy optimisation} (MPO) is used to fit $Q$-function and policy $\pi$ based on uniformly drawn samples from the policy replay.
The resulting policy $\pi^{i+1}$ is then synced back into the actor.
Note that both model and policy learning is executed in independent loops.
}
\label{fig:method-diagram}
\end{center}
\vskip -0.2in
\end{figure*}

\section{Related Work}

The utilisation of forward modelling error as reward signal has been implemented in deterministic and probabilistic settings~\citep{shelhamer2016loss_is_reward,achiam2017surprise} and is commonly known as \emph{curiosity learning}.
The error reward signal encourages the exploration of unfamiliar parts of the state-action space which are not yet well-predictable.
\cite{pathak2017curiosity} derive the curiosity reward from an inverse dynamics model which is simultaneously less prone to be confused by unpredictable elements of the environment (cf. \emph{white-noise problem}~\citep{schmidhuber2010formal_creativity}).
\cite{burda2018rnd} use a randomly initialised projection from observation into latent space and the predictor driving the learning process is tasked with learning this projection.
\cite{huang2019penalty_surprise} use the error of a predictive model of penalties in dexterous manipulation and find that the additional intrinsically motivated exploration helps in the development of gentle grasping policies.
Curiosity learning can also be seen through an information theoretic lens using an information gain objective like in \cite{still2012it_curiosity_rl} or \cite{houthooft2016vime}.
Lastly, the \emph{disagreement} between an ensemble of forward models~\citep{pathak2019disagreement} can also be treated as a proxy for model uncertainty and exploited as a reward signal.
In addition to serving as a reward generator, the predictors can also be used for targeted exploration.
\cite{lowrey2018plan} and \cite{sekar2020planning_to_explore} employ world models to deliberately explore in regions where the expected prediction error is high \emph{a priori} - as opposed to realising that something surprising has been observed \emph{a posteriori} like in standard curiosity approaches.

Besides curiosity learning, the literature features an extensive body of work on structured exploration methods in reinforcement learning.
Classical methods like count-based exploration schemes have been revisited in continuous settings, e.g. in~\cite{bellemare2016unifying} changes in state density estimations are used as exploration bonus rewards.
Another line of work revolves around the central notion of \emph{empowerment}~\citep{klyubin2005empowerment} which aims to find new behaviours which are increasingly controllable by the agent~\citep{mohamed2015variational,gregor2016variational}.
The \emph{estimation of model learning progress} has also been discussed as a reward signal for exploration but this approach is much harder to implement as it relies on a measure of model improvement~\citep{lopes2012exploration}.
Closely related to model-improvement-based exploration is the idea of \emph{PowerPlay}~\citep{schmidhuber2013powerplay} which describes an inductive task proposal scheme equipping the agent with a mechanism which infers the current frontier of soluble tasks and inductively creates a novel task which can be solved by employing the agent's current knowledge.
However, this exploration scheme has mostly remained conceptual so far with experiments limited to simple pattern recognition tasks~\citep{srivastava2013first_powerplay}.
Related to PowerPlay's idea of task proposition but supposedly more tractable is the concept of \emph{self-play}~\citep{sukhbaatar2017selfplay_curriculum}.
In this paradigm, two agents play a competitive game where one player is rewarded for inventing a behaviour which the other agent cannot imitate.
This approach has recently led to the emergence of highly complex object manipulation behaviour on a simulated robotic arm~\citep{openai2021asymmetric_selfplay}.

The importance of diverse interactions with the environment -- similar to the playful behaviour exhibited by children -- has been stressed in recent works in reinforcement learning, e.g. in~\cite{haber2018emergence} or \cite{lynch2020lfp}.
Such intrinsically motivated exploration has been shown to lead to the discovery of diverse behaviour and re-usable skills as a natural `by-product' of interacting with the environment~\citep{singh2005intrinsically}.
Several recent works have identified the \emph{diversity of behaviour} as a central objective to optimise for during unsupervised exploration~\citep{eysenbach2018diayn,sharma2019dads}.
\cite{sharma2020emergent} have extended this line of work and also showed that the acquired latent `skill space' can be leveraged in model-predictive control fashion for goal-conditioned navigation on a real-world quadruped.

Our work builds upon the curiosity learning approach utilising the forward prediction error of a dynamics model as a reward signal.
However, in contrast to typical curiosity setups~\citep{burda2018curiosity_study} which are optimised \emph{on-policy} we employ an \emph{off-policy} method to train the agent.
Our method is also set apart from prior art with regards to the utilisation of self-discovered behaviour.
Instead of using \emph{model-predictive control}~\citep{sharma2020emergent}, we leverage emergent behaviour directly by employing policy snapshots as modular skills in a mixture policy~\citep{wulfmeier2019rhpo,wulfmeier2021data}.

\section{Method}
\label{sec:method}

In this section, we present SelMo -- a self-motivated exploration method which optimises a curiosity objective in an off-policy fashion.
Our system is designed around two key components: A forward dynamics model $f_{dyn} \colon S \times A \mapsto S$ which aims to approximate the state transition function of the environment and a policy $\pi(a_t \vert s_t)$ which aims to take transitions in the environment for which the prediction error of $f_{dyn}$ is high.

Given potentially different requirements for the dynamics model and the exploration policy, the implementation builds on two separate learning processes.
We have decided to implement the system in this distributed way to more easily realise the data labeling process and have convenient control over learning rate and data flow parameters.
However, other implementations which fuse both learning processes and data buffers are also conceivable.
Crucially, our setup deviates from recent on-policy approaches for curiosity-based rewards in two important aspects:
First, we optimise in an off-policy fashion based on a diverse set of experienced transitions in the environment.
In this way, the policy is encouraged to to cover a broader spectrum of exploration avenues as the learning evolves.
Second, we employ an approximate but efficient data labelling strategy which only assigns curiosity rewards to trajectories when they are used to update the dynamics model, but refrain from relabeling all trajectories in the policy replay after every model update.
We provide a general overview over the whole system in \cref{fig:method-diagram} and explain each component in detail in the paragraphs below.

\paragraph{Model Replay}
The fixed-size model replay buffer $D_\mathcal{M}$ stores trajectories $\tau^k = \left[ (s^k_0, a^k_0, s^k_1), \dots, (s^k_{T-1}, a^k_{T-1}, s^k_T) \right]$ collected by the actor in the environment.
The values of all environment transitions $(s_t, a_t, s_{t+1})$ are normalised in the range $\left[-1, 1\right]$.
Batches of training data for the world model are sampled uniformly from this buffer as $\mathcal{B}_\mathcal{M} := \{\tau^1, \dots, \tau^B\} \sim \mathcal{U}(D_\mathcal{M})$.
In order to preserve a diverse sampling of environment transitions during the whole learning process, each trajectory in $D_\mathcal{M}$ can be sampled at most $n_{max}$ times.
Additionally, old trajectories in the buffer are replaced by new ones at random when the buffer size limit is exceeded.

\paragraph{World Model}
We describe the environment in which the agent operates as $\mathcal{E} = (S, A, P)$ with state and action spaces $S$ and $A$ as well as a state transition function $s_{t+1} = P(s_t, a_t)$ over discrete time steps which describes the environment's \emph{transition dynamics}.
The world model is a forward-predictive model $f_{dyn} \colon S \times A \mapsto S$ which approximates the environment's transition dynamics as:
\begin{equation}
    \hat{s}_{t+1} = f_{dyn}(s_t, a_t; \theta)
    \label{eq:world_model}
\end{equation}
In our case, \cref{eq:world_model} is implemented as a two-layer MLP with parameters $\theta$.
Besides estimating the transition dynamics from observed data, the world model plays a crucial role in assigning the reward for each observed transition $(s_t, a_t, s_{t+1})$.
When a transition is evaluated by the world model, the assigned reward is scaled by the model's current prediction error.
\begin{equation}
    r^{(j)}(s_t, a_t, s_{t+1}) = \tanh(\eta_r * (f_{dyn}^{(j)}(s_t, a_t) - s_{t+1})^2)
    \label{eq:reward-curiosity}
\end{equation}
The `state' of the world model is indicated by the number of gradient updates $j$ which have been performed on it so far.
We scale the reward via a hyper-parameter $\eta_r$ and pass it through a $\tanh$ to keep it bounded for the downstream policy learning procedure.
When a new batch of data $\mathcal{B}$ is sampled from the model replay, the world model performs two operations.
First, it labels each $\tau \in \mathcal{B}$ by assigning curiosity rewards $r^C$ according to \cref{eq:reward-curiosity}.
Second, it performs one gradient update $\theta^{(j+1)} \leftarrow \theta^{(j)} + \eta_\mathcal{M} \partial \mathcal{L}_{dyn}^{(j)}(\mathcal{B}) / \partial \theta^{(j)}$ by minimising its prediction loss:
\begin{equation}
    \mathcal{L}_{dyn}^{(j)}(\mathcal{B}) = \sum_{\tau \in \mathcal{B}} \sum_{(s_t, a_t, s_{t+1}) \in \tau}(f_{dyn}^{(j)}(s_t, a_t) - s_{t+1})^2
    \label{eq:loss-world_model}
\end{equation}
After one world model update, the relabeled batch of trajectories $\tilde{\mathcal{B}}$ is stored in the policy replay buffer $D_\pi$.

\paragraph{Policy Replay}
The fixed-size policy replay $D_\pi$ stores tuples $(\tau^j, r^j)$ representing trajectories which have been labeled with curiosity rewards by the world model.
This off-policy setup provides the policy learner with a diverse training set to optimise for useful exploration actions \emph{globally} and not only in the vicinity of the most recent experience.
During policy learning, data batches are sampled uniformly from this buffer as $\mathcal{B}_\pi := \{(\tau^1, r^1), \dots, (\tau^B, r^B)\} \sim \mathcal{U}(D_\pi)$.
Similar to the model replay, each tuple can be sampled up to  $m_{max}$ times to increase the utilisation of each data point during policy learning.
However, the removal strategy of the policy replay is $FIFO$ to ensure that trajectories with the most outdated curiosity rewards get replaced first to reflect the change in the world model, albeit with a certain delay.

\paragraph{Policy}
The \emph{Markov Decision Process} (MDP) which is induced by this setup can be written as:
$\mathcal{M}^{(j)} = (S, A, P, r^{(j)})$.
Since the world model $f_{dyn}^{(j)}$ changes with every gradient update, the reward function $r^{(j)}$ changes continuously.
Consequently, the reward varies with the model training timesteps $j$ and the policy $\pi$ is required to keep adapting.
The policy replay $D_\pi$ is filled by the world model's training loop. It contains data of the most recent $\kappa = \vert D_\pi \vert / \vert \mathcal{B} \vert$ versions of the MDP.
Hence, the resulting policy is optimising for a mixture of MDPs $\{(S,A,P,r^{(\iota)} \mid \iota \in [j-\kappa, \dots, j]\}$.
Policy and critic network are implemented as two separate MLPs.
The policy is optimised off-policy using MPO~\citep{abdolmaleki2018mpo} and a separate learning rate $\eta_\pi$.

\section{Experiments}
This section is split in two parts.
First, we report our analysis of the emergence of behaviours when optimising only for a curiosity reward in \cref{sec:exp-emergence}.
Second, we present an empirical utilisation of self-discovered behaviour for accelerated learning of new downstream tasks in \cref{sec:exp-utilisation}.
For details regarding the simulation domains and model hyper-parameters, we refer the reader to \cref{sec:exp-domains} and \cref{sec:exp-hparams} respectively.

\subsection{Emergence of Behaviour}
\label{sec:exp-emergence}
We start our investigation with an analysis of the behaviour which emerges in our two simulated domains as depicted in~\cref{fig:timeline-qualitative}.
The JACO domain features a 9 DoF robotic arm and two cubes; the OP3 domain features a 20 DoF humanoid robot.
For this experiment, we run the SelMo learning loop (cf.~\cref{fig:method-diagram}) with a single actor for 100K episodes on each of the environments.
During training, the agent solely optimises its curiosity objective which is defined by~\cref{eq:reward-curiosity}.
The visual inspection of the experiments reveal that in both cases, diverse sets of human-interpretable behaviour emerge consistently and are exhibited by the agent for extended periods of time before the ever-changing curiosity reward function shifts the learning towards a new behaviour.
Below, we describe the observed behaviours in each domain in greater detail.

\begin{figure*}[t]
\vskip 0.2in
\begin{center}
\centerline{\includegraphics[width=\textwidth]{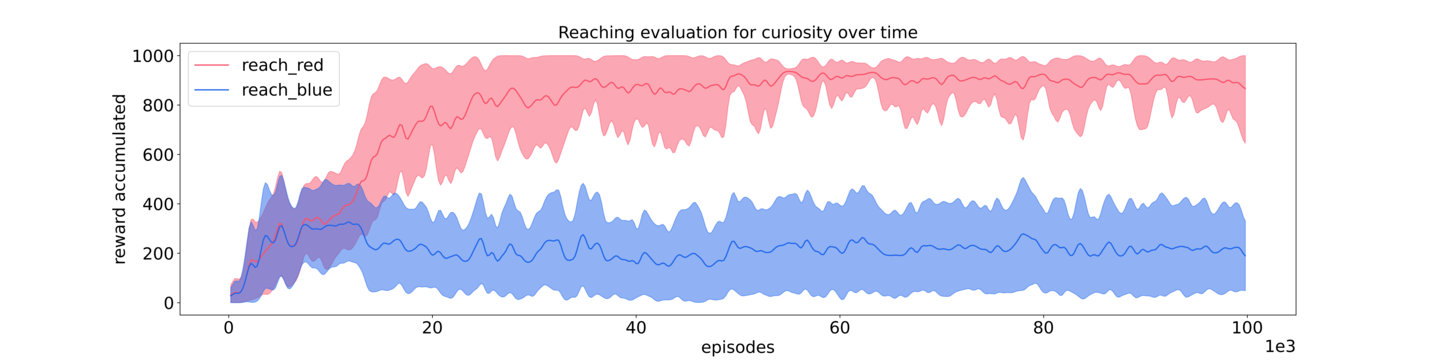}}
\centerline{\includegraphics[width=\textwidth]{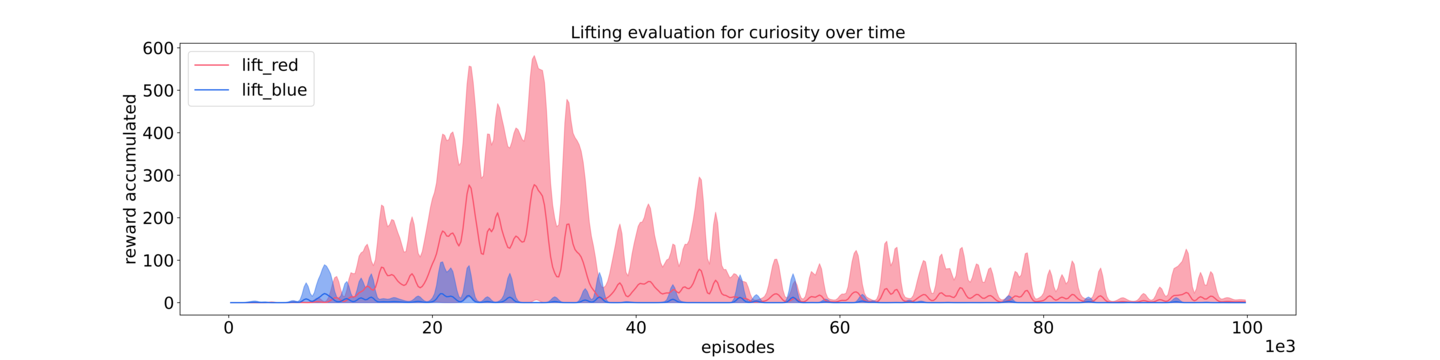}}
\caption{
Manipulation task evaluation in the JACO environment over the lifetime of one experiment while the agent is only trained on the curiosity objective (cf.~\cref{eq:reward-curiosity}).
A snapshot of the curiosity policy is saved every 100 episodes and evaluated on reaching and lifting the red and blue cubes respectively.
Mean and standard deviation are plotted over 20 evaluation runs per policy snapshot and the plot is smoothed with an exponential filter of $\sigma=1.5$.
}
\label{fig:timeline-quantitative-jaco}
\end{center}
\vskip -0.2in
\end{figure*}

\subsubsection*{Emergent Manipulation Behaviour on JACO} 
A qualitative example timeline of emerging behaviours on the JACO arm is depicted in \cref{fig:timeline-qualitative} (top) and supplemented by a plot evaluating reaching and lifting behaviour during the run in \cref{fig:timeline-quantitative-jaco}.
We find that the agent is very quickly driven towards both cubes with equal attention and starts interacting with them by pushing them around.
Soon thereafter, it discovers that pushing them up the slanted walls of the bin facilitates picking them up before it stably latches onto a mode in which it prefers manipulating the red cube over the blue one after approximately 15K episodes.
This also coincides with a first period of sustained lifting of the red cube.
We hypothesise that the discovery of lifting said cube reinforces the interaction with it as it opens up a new dimension along which model prediction error can be rewarded: the height of the object.
After about 25K episodes, it has learned to pick up an object reliably even without the help of the slopes.

After approximately 40K episodes, the curiosity objective pushes the agent to deliberately take objects outside of the workspace and perform pick-and-place operations which move a cube over a long distance but at a lower height.
Interestingly, the policy does not degenerate into extreme behaviours like spinning motions which have been observed in related work~\citep{sekar2020planning_to_explore} but stays focused on the objects and keeps exploring their physical properties.
For instance, at around 70K episodes, the policy investigates the stability of cube poses in a targeted way by balancing them on their edges and corners.
Finally, after about 80K episodes, it starts exploring the possibilities of moving both cubes simultaneously.

\begin{figure*}[t]
\begin{center}
\centerline{\includegraphics[width=\textwidth]{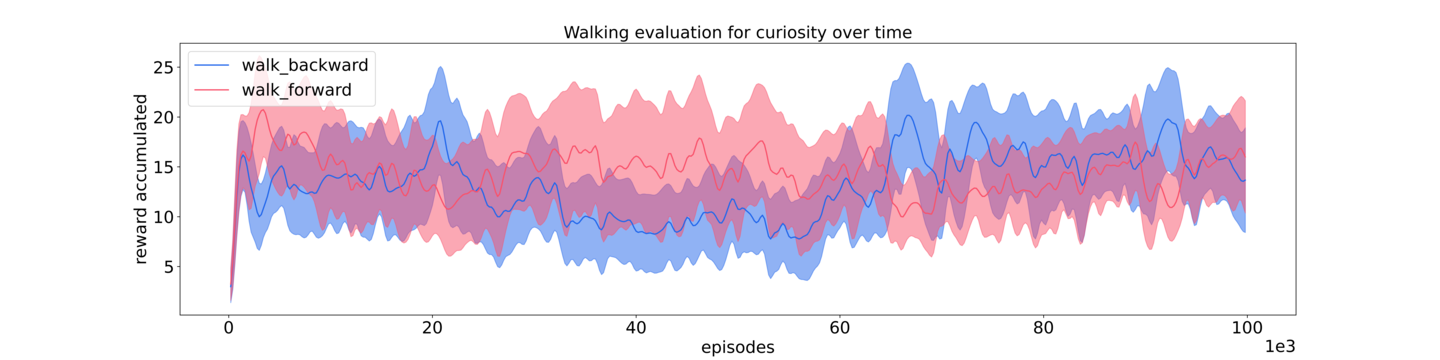}}
\caption{
Task evaluation in the OP3 environment over the lifetime of one experiment while the agent is only trained on the curiosity objective (cf.~\cref{eq:reward-curiosity}).
A snapshot of the curiosity policy is saved every 100 episodes and evaluated on locomotion (\texttt{walk\_\{forward,backward\}}) tasks.
Mean and standard deviation are plotted over 20 evaluation runs per policy snapshot and the plot is smoothed with an exponential filter of $\sigma=1.5$.
}
\label{fig:timeline-quantitative-op3}
\end{center}
\vskip -0.2in
\end{figure*}

\subsubsection*{Emergent Locomotion Behaviour on OP3}
Similar to the JACO arm, we present a timeline of emerging behaviour on the OP3 in \cref{fig:timeline-qualitative} (bottom) and a corresponding evaluation of locomotion behaviour in \cref{fig:timeline-quantitative-op3}.
Unsurprisingly, the agent spends roughly the first 2K episodes -- indicated by the steep rise in walking rewards -- just on learning a sense of balance because an episode is terminated early when the torso constraint (cf.~\cref{sec:exp-domains}) is violated and the agent is about to fall.
This also corresponds to maximising the experienced episode length because this increases the chances of further increasing the accumulated reward.
This finding is in line with earlier work~\citep{pathak2017curiosity} which has shown that the avoidance of a `death' event is a natural by-product of curiosity-driven learning with a positive reward and favourably shapes the emerging policy.

Once the agent has learned to stay upright, it slowly starts to develop basic locomotion in the form of stumbling forwards and backwards with only small foot lifting heights which is also reflected in minor oscillations during the evaluation of the walking rewards in \cref{fig:timeline-quantitative-op3}.
Interestingly, after around 30K episodes, the agent has discovered to swing its arms to take bigger steps.
This is most impressively first demonstrated after nearly 40K episodes when the agent balances on one foot while stretching out the other leg using its arms for counterbalancing moves.
Using the arms also opens new avenues for exploration.
Approximately 40K episodes into training, the agent has learned to catch itself when falling backwards.
This leads to the discovery of a sit-down behaviour which does not violate the environment's torso constraints.
After the agent has explored various `ground exercises' it switches back to walking gaits at around 55K episodes.
Then, the whole body movement has become considerably more nimble and its movement repertoire now features quick turns, stumbling reflexes and even safe backward leaps.
After about 70K episodes the agent starts revisiting earlier behaviour, e.g. the balancing skill, but keeps adding variations to it like knee-bending or stretching.

\subsection{Utilisation of Emergent Behaviour}
\label{sec:exp-utilisation}
As we have discussed in the previous section, the constantly evolving curiosity policy develops behaviours which correspond to the solution of concrete tasks (cf.~\cref{fig:timeline-quantitative-jaco},~\cref{fig:timeline-quantitative-op3}).
In order to retain those diverse behaviours to accelerate the learning of new tasks, we have devised the following experiment:
Assuming that the curiosity policies exhibit undirected yet versatile exploratory behaviour, we investigate how well they could serve as auxiliary skills in a modular learning setup of a downstream task.

We employ \emph{Regularized Hierarchical Policy Optimization} (RHPO)~\citep{wulfmeier2019rhpo} as this framework allows us to compose multiple policies in a hierarchical manner.
In each environment, we define a downstream target task which we are interested in learning and provide five policy snapshots from a SelMo experiment as auxiliary exploration skills.
During the SelMo experiment, we optimise solely for the curiosity objective and save a snapshot of the curiosity policy every 100 episodes.
The RHPO experiment subsequently samples SelMo policy snapshots and utilises the behaviour exhibited by them to assist the exploration for the desired downstream task.

\begin{figure*}[t]
\begin{center}
\centerline{\includegraphics[width=\textwidth]{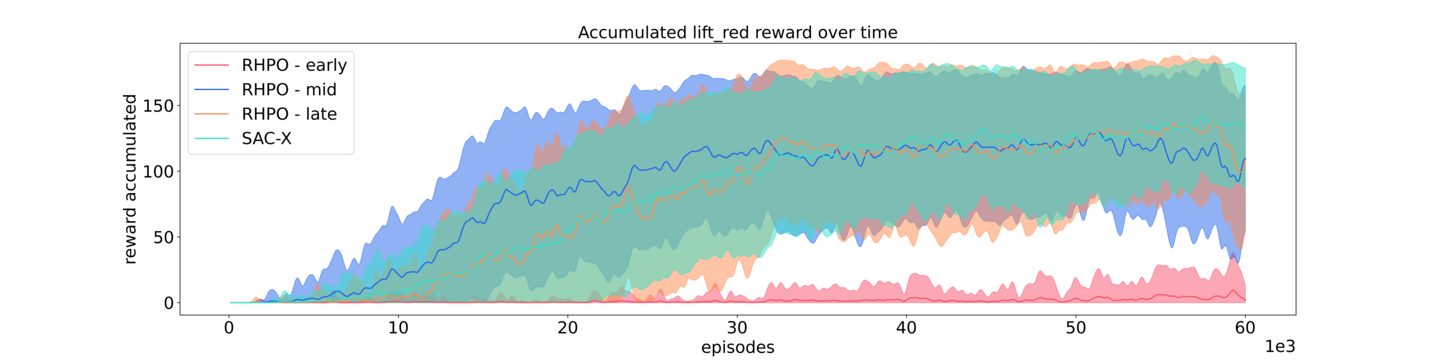}}
\caption{
Learning curves for hierarchical skill learning of \texttt{lift\_red} using SelMo auxiliary policies in the JACO environment.
The SAC-X baseline uses auxiliary reward functions \texttt{\{reach,move\}\_red}.
Each RHPO run uses five randomly sampled SelMo policies from the respective intervals as auxiliary skills (cf.~\cref{sec:exp-utilisation}).
Mean and standard deviation are plotted for five random seeds for each model and the plot is smoothed with an exponential filter of $\sigma=1.5$.
}
\label{fig:rhpo-learning}
\end{center}
\vskip -0.2in
\end{figure*}

For this experiment, we refer again to the JACO environment where we define the target task to be \texttt{lift\_red}.
While the policy for the target task is randomly initialised, the auxiliary policies are randomly sampled from the SelMo snapshots and kept fixed during the entire RHPO training.
We differentiate between three different phases from which the SelMo snapshots are chosen: \texttt{early} comprises of snapshots which have been trained on up to 10K episodes, \texttt{mid} snapshots are from the interval between 10K and 20K episodes and \texttt{late} snapshots are sampled between 20K and 30K training episodes.
In~\cref{fig:rhpo-learning} we compare the learning progress of the downstream task with the sampled SelMo auxiliary skills against a baseline featuring hand-designed task curricula in an SAC-X framework~\citep{riedmiller2018sacx}.
In the case of the JACO environment, the agent is given a curriculum of reward functions which help to \texttt{reach} and \texttt{move} the \texttt{red} cube.

In the case of \texttt{lift\_red} on JACO, we find that SelMo auxiliaries from the \texttt{mid} and \texttt{late} exploration periods give the learning of the lifting policy a significant boost which is commensurate with a tuned SAC-X baseline featuring multiple auxiliary rewards which have been hand-designed to facilitate the learning of \texttt{lift\_red}.
This result is in line with the performance observed in \cref{fig:timeline-quantitative-jaco} where the curiosity-based policy has developed a sustained lifting behaviour between 20K and 35K episodes.
Consequently, snapshot auxiliaries sampled from that range are particularly useful when a targeted lifting policy is to be learned.

This experiment shows that even a simple behaviour retention strategy like policy snapshotting can already provide clear benefits for downstream learning of new tasks.
The self-discovered behaviours from a curiosity training phase afford a task learning scaffold which can be commensurate with a specifically designed set of auxiliary reward functions.
This is a promising result suggesting that independent curious exploration could be used in lieu of human-engineered task curricula in complex manipulation scenarios.

\section{Discussion}
Our experiments have shown that complex manipulation and locomotion behaviour such as grasping, lifting, balancing, sitting and leaping emerges completely unsupervised in an off-policy curiosity learning setup on a 9 DoF robot arm and a 20 DoF humanoid.
This observation supports our hypothesis that self-discovered behaviour can provide a valuable skill repertoire for the learning of new downstream tasks.
We add further evidence to this hypothesis by utilising randomly selected policy snapshots from a curiosity training as auxiliary skills in a modular learning setup and show that they provide a learning scaffold commensurate with hand-designed auxiliary reward functions for the respective tasks.
This suggests that curiosity-based exploration should be treated as an independent aspect of a learning system as opposed to a mere bonus reward or policy pre-training phase.
In this study, we provide a baseline for harnessing self-discovered behaviour by randomly sampling policy snapshots and treating them as fixed skills in a modular learning setup.
However, more sophisticated techniques for identification, retention and utilisation of behaviour in curious exploration settings are conceivable and will be briefly touched upon in this discussion section.

\paragraph{Identification of Emerging Behaviour}
As we have shown in~\cref{sec:exp-emergence}, complex behaviour can emerge in an unsupervised way when optimising a policy for a curiosity objective.
However, it is revealed in \cref{fig:timeline-quantitative-jaco} and \cref{fig:timeline-quantitative-op3} that pre-conceived reward functions are only able to capture a fraction of the emergent behaviour like reaching and lifting of individual objects or basic walking gaits.
More involved behaviours are not covered by basic reward functions and implementing reward functions to identify a broad set of behaviour a priori does not scale well with the unsupervised nature of curiosity learning.
Diversity-based approaches~\citep{eysenbach2018diayn,sharma2019dads,sharma2020emergent} address this challenge via a latent `skill space' which modulates the policy network to capture different modes of operation like jumping and walking.
The temporal and hierarchical abstraction provided by the latent vector also facilitates planning over the skill space.
However, the identification of particularly useful skills worth retaining and comparatively useless skills which could be overwritten is still an open question, especially in never-ending learning settings where a curiosity-driven exploration keeps discovering new behaviours or revisiting old ones.

\paragraph{Retention and Utilisation of Self-Discovered Behaviour}
Closely related to the question of behaviour identification is the question of its retention and utilisation.
In our setup we have treated snapshots of the same curiosity policy as different skills as they represent different behaviour over time.
Using a latent skill space as an arbiter for different behaviours in the same policy network~\citep{eysenbach2018diayn,sharma2019dads} has also been shown to work effectively as this formulation also enables planning over the learned space of skills.
However, \cite{lynch2020lfp} have also demonstrated the utility of raw \emph{play data} in the training of versatile goal-conditioned policies which would position the curiosity-driven exploration as a data collector for a downstream policy distillation instead of a re-usable behaviour in itself.
Lastly, \cite{riedmiller2018sacx} and \cite{hertweck2020ssi} have shown the benefits of curricula of reward functions for the learning of complex manipulation tasks which opens another avenue for the utilisation of curiosity-based learning:
Instead of using frozen snapshots of the policy as a repertoire of skills one could also treat different versions of the world model as a set of distinct reward functions to encourage the optimisation for diverse behaviour during learning of new downstream tasks.

\section{Conclusion}
In this paper we have studied the emerging behaviour when optimising an exploration policy for a curiosity objective derived from a forward-predictive world model.
To this end, we have presented SelMo, a curiosity-based, off-policy exploration method and applied it in two continuous control domains: a simulated robotic arm and humanoid robot.
We have observed that complex behaviour emerges in both settings and provided a baseline for the utilisation of this self-discovered behaviour in a modular downstream learning scenario.
Despite the remaining technical challenges, we believe that the automatic identification and retention of useful emerging behaviour from curious exploration is a fruitful avenue of future investigation in unsupervised reinforcement learning.

\section*{Acknowledgements}
We would like to thank Arunkumar Byravan, Dushyant Rao, Abbas Abdolmaleki, Yuval Tassa and Nathan Lambert for the discussions and feedback during the conception and execution of this project.
We would also like to thank Andrea Huber for facilitating all organisational aspects of this collaboration.

\section*{Author Contributions}
\textbf{Oliver Groth} developed and implemented the SelMo model, conducted the curiosity experiments, studied related work on self-motivated reinforcement learning and wrote the paper.
\textbf{Markus Wulfmeier} advised during the implementation and experimentation phase and helped writing the paper.
\textbf{Giulia Vezzani} conducted the RHPO experiments and created the experiment evaluation plots.
\textbf{Vibhavari Dasagi} assisted in compiling related work and provided feedback on the method and discussion section.
\textbf{Tim  Hertweck} provided technical support during the implementation and experimentation phase and created the supplementary videos.
\textbf{Roland Hafner} conducted the SAC-X baseline experiments and provided support in setting up the JACO and OP3 environments.
\textbf{Nicolas Heess} provided feedback on the experimental results and during writing and revision of the paper.
\textbf{Martin Riedmiller} conceived the idea of self-motivated skill learning, provided conceptual feedback and supervised the project.

\bibliographystyle{abbrvnat}
\setlength{\bibsep}{5pt} 
\setlength{\bibhang}{0pt}
\bibliography{selmo}

\clearpage
\appendix
\section{Simulation Environments}
\label{sec:exp-domains}

In this section we provide details about the two robotic simulation domains used in our experiments.

\begin{figure}[h!]
\begin{center}
\centerline{\includegraphics{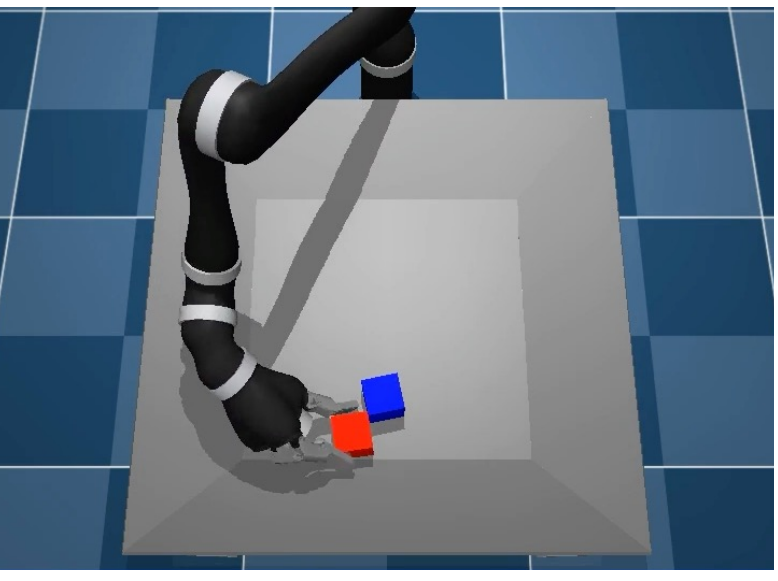}}
\caption{The JACO manipulation environment. The 6 DoF robot arm with a 3 DoF gripper can interact with multiple same-sized cubes in its workspace.}
\label{fig:env-jaco}
\end{center}
\vskip -0.2in
\end{figure}

\begin{table}[h!]
\caption{State and action space semantics of JACO environment.}
\label{tab:jaco-state}
\vskip 0.15in
\begin{center}
\begin{small}
\begin{tabular}{lr}
\toprule
\textsc{Feature} & \textsc{Dimension} \\
\midrule
\texttt{arm/joints\_pos} & $6$ \\
\texttt{arm/joints\_vel} & $6$ \\
\texttt{arm/hand/finger\_joints\_pos} & $3$ \\
\texttt{arm/hand/finger\_joints\_vel} & $3$ \\
\texttt{arm/hand/fingertip\_sensors} & $3$ \\
\texttt{arm/hand/pinch\_site\_pos} & $3$ \\
\textbf{proprioception} & $\sum = 24$ \\
\midrule
\texttt{O<i>/rel\_pos\_wrt\_tcp} & $3$ \\
\texttt{O<i>/pos} & $3$ \\
\texttt{O<i>/orientation} & $4$ \\
\texttt{O<i>/linvel} & $3$ \\
\texttt{O<i>/angvel} & $3$ \\
\texttt{O<i>/proptype} & $5$ \\
\texttt{O<i>/dimensions} & $3$ \\
\textbf{perception per object <i>} & $\sum = 24$ \\
\midrule
\texttt{arm} & $6$ \\
\texttt{hand} & $3$ \\
\textbf{action space} & $\sum = ~9$ \\
\bottomrule
\end{tabular}
\end{small}
\end{center}
\vskip -0.1in
\end{table}

\subsection{JACO Manipulation Environment}
This environment is designed to study manipulation tasks like object lifting and stacking with a robotic arm (cf. \cref{fig:env-jaco}).
The state observation space consists of 72 dimensions: 24 features are used to represent the robot's proprioception as well as the state of each object.
The action space spans 9 dimensions: 6 concerning the arm and 3 concerning the three-point gripper.
A detailed description of the environment is provided in \cref{tab:jaco-state}.
Interactions with the two objects (\texttt{O1} = red cube, \texttt{O2} = blue cube) are evaluated using the sparse reward functions \texttt{reach\_\{red,blue\}} and \texttt{lift\_\{red,blue\}}.
Each episode in this environment lasts 20 seconds or 400 control timesteps.

\begin{figure}[h!]
\begin{center}
\centerline{\includegraphics{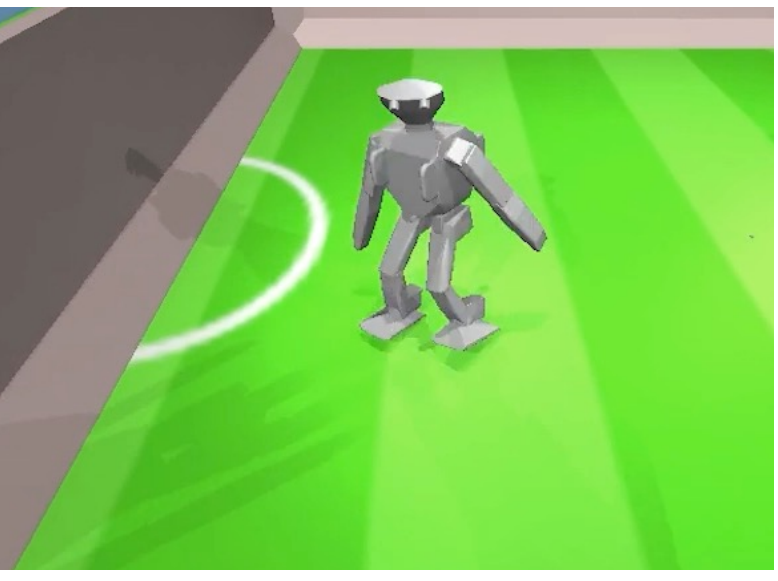}}
\caption{The OP3 locomotion environment. The 20 DoF humanoid robot can walk around on a plane. The episode terminates early, if the OP3 is about to fall over.}
\label{fig:env-op3}
\end{center}
\vskip -0.2in
\end{figure}

\begin{table}[h!]
\caption{State and action space semantics of OP3 environment.}
\label{tab:op3-state}
\vskip 0.15in
\begin{center}
\begin{small}
\begin{tabular}{lr}
\toprule
\textsc{Feature} & \textsc{Dimension} \\
\midrule
\texttt{walker/joints/pos} & $20$ \\
\texttt{walker/imu/linear\_acc} & $3$ \\
\texttt{walker/imu/angular\_vel} & $3$ \\
\texttt{walker/imu/gravity} & $3$ \\
\texttt{scaled/action\_filter/state} & $20$ \\
\textbf{proprioception} & $\sum = 49$ \\
\midrule
\texttt{head\_\{pan,tilt\}} & $2$ \\
\texttt{\{l,r\}\_ankle\_\{pitch,roll\}} & $4$ \\
\texttt{\{l,r\}\_elbow} & $2$ \\
\texttt{\{l,r\}\_hip\_\{pitch,roll,yaw\}} & $6$ \\
\texttt{\{l,r\}\_knee} & $2$ \\
\texttt{\{l,r\}\_shoulder\_\{pitch,roll\}} & $4$ \\
\textbf{action space} & $\sum = 20$ \\
\bottomrule
\end{tabular}
\end{small}
\end{center}
\vskip -0.1in
\end{table}

\subsection{OP3 Locomotion Environment}
This environment is designed to study locomotion with a humanoid robot (cf. \cref{fig:env-op3}).
The state observation consists of 49 proprioceptive features.
The 20-dimensional action space controls the orientations of the robot's head, ankle, elbow, hip, knee and shoulder.
Actions passed to the robot are smoothed with an exponential filter to reduce motion jerk.
A detailed description of the environment is provided in \cref{tab:op3-state}.
The robot always spawns in a standing, upright position.
Locomotion is evaluated by the dense reward functions \texttt{walk\_\{forward,backward\}} for forward and backward walking gaits respectively.
Each episode in this environment lasts 10 seconds or 200 control timesteps.
If the robot's hip angle deviates more than $15^\circ$ from an upright orientation, the simulation terminates early effectively preventing the robot from falling over.

\section{Model and Training Details}
\label{sec:exp-hparams}
Across all experiments with the SelMo architecture, we consistently use the following hyper-parameters and model architectures.
Each SelMo experiment is run with a single actor for $N = 1e5$ episodes.

\paragraph{World Model $f_{dyn}$}
The world model is implemented as a \emph{multi-layer perceptron} (MLP) with the following layer sizes and activation functions:
\texttt{[FC(256), elu($\cdot$), FC(256), elu($\cdot$), FC(size\_state)]}
For each environment, \texttt{size\_state} is the sum of the dimensions for proprioception and perception (cf.~\cref{sec:exp-domains}).
The world model is optimised using Adam~\citep{kingma2014adam} with a learning rate of $\eta_\mathcal{M} = 3e-4$.

\paragraph{Policy}
Both policy $\pi$ and critic $Q$ are implemented as two independent MLPs with the following layer sizes and activation functions:
\begin{itemize}
    \item $Q$: \texttt{[tanh($\cdot$), FC(512), elu($\cdot$), FC(512), elu($\cdot$), FC(256), FC(1)]}
    \item $\pi$: \texttt{[FC(256), elu($\cdot$), FC(256), elu($\cdot$), FC(128), FC(size\_action)]}
\end{itemize}
The \texttt{size\_action} is different for each environment (cf.~\cref{sec:exp-domains}).
The policy is optimised using Adam~\citep{kingma2014adam} with a learning rate of $\eta_\pi = 3e-4$.
The reward scale is set to $\eta_r = 10.0$ across all experiments.

\paragraph{Replays}
The replays $D_\mathcal{M}$ and $D_\pi$ store trajectories with a length of $T = 50$ transitions.
The buffer sizes used are $|D_\mathcal{M}| = |D_\pi| = 5e4$ and each trajectory in the buffers can be sampled at most $n_{max} = m_{max} = 32$ times (cf.~\cref{sec:method}).
The batch size of samples drawn from the replays is set to $B = 64$.

\end{document}